\documentclass[10pt,twocolumn,letterpaper]{article}

\usepackage[pagenumbers]{cvpr} 

\usepackage{multirow}
\usepackage{kotex}
\usepackage[most]{tcolorbox}
\newtcolorbox{promptblock}{
  enhanced, breakable,
  colback=gray!3, colframe=black!20,
  boxrule=0.5pt, arc=2pt,
  left=6pt,right=6pt,top=6pt,bottom=6pt,
  fontupper=\ttfamily\small
}

\definecolor{cvprblue}{rgb}{0.21,0.49,0.74}
\usepackage[pagebackref,breaklinks,colorlinks,allcolors=cvprblue]{hyperref}

\def\paperID{*****} 
\def\confName{CVPR}
\def\confYear{2026}

\title{Integrating Multimodal Large Language Model Knowledge into Amodal Completion}

\author{
Heecheol Yun$^{1}$ \quad
Eunho Yang$^{1,2}$ \\
$^{1}$KAIST \quad
$^{2}$AITRICS \\
{\tt\small \{yoon6503, eunhoy\}@kaist.ac.kr}
}

\begin{document}
\twocolumn[{%
\renewcommand\twocolumn[1][]{#1}%
\maketitle
\includegraphics[width=\linewidth]{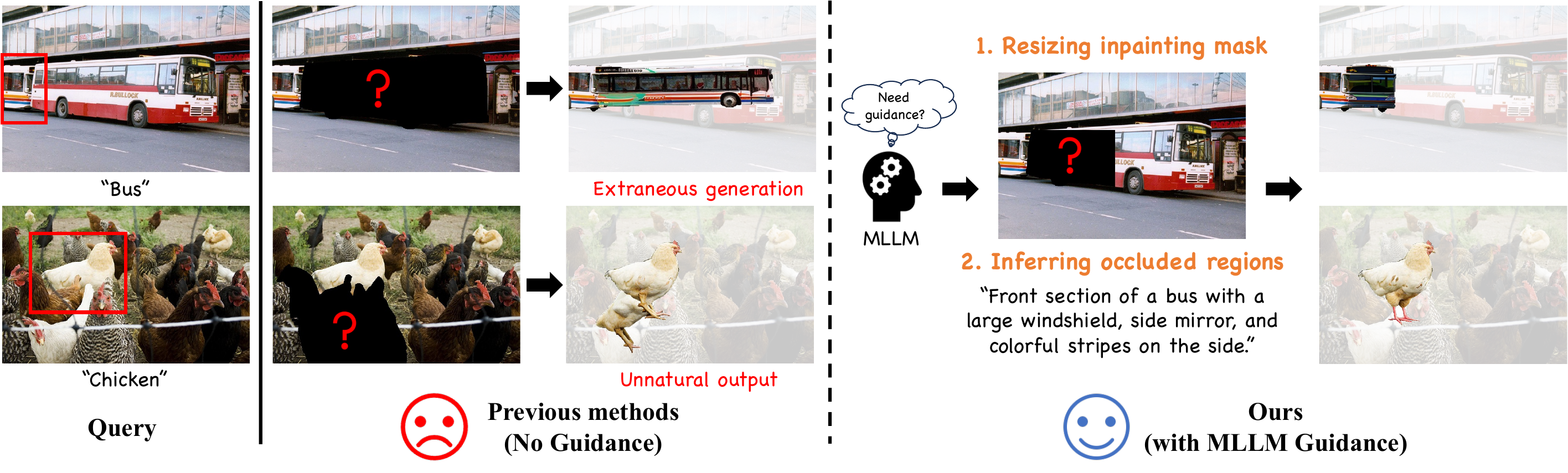}
\captionof{figure}{Our method selectively leverages MLLMs to reason about the extent and content of occluded parts. Incorporating them into amodal completion effectively enhances performance. \vspace{1em}}
\label{fig:teaser}
}]
\begin{abstract}
With the widespread adoption of autonomous vehicles and robotics, amodal completion, which reconstructs the occluded parts of people and objects in an image, has become increasingly crucial. Just as humans infer hidden regions based on prior experience and common sense, this task inherently requires physical knowledge about real-world entities. However, existing approaches either depend solely on the image generation ability of visual generative models, which lack such knowledge, or leverage it only during the segmentation stage, preventing it from explicitly guiding the completion process. To address this, we propose \textit{AmodalCG}, a novel framework that harnesses the real-world knowledge of Multimodal Large Language Models (MLLMs) to guide amodal completion. Our framework first assesses the extent of occlusion to selectively invoke MLLM guidance only when the target object is heavily occluded. If guidance is required, the framework further incorporates MLLMs to reason about both the (1) \textit{extent} and (2) \textit{content} of the missing regions. Finally, a visual generative model integrates these guidance and iteratively refines imperfect completions that may arise from inaccurate MLLM guidance. Experimental results on various real-world images show impressive improvements compared to all existing works, suggesting MLLMs as a promising direction for addressing challenging amodal completion.
\end{abstract}
    
\section{Introduction}
\label{sec:intro}

Imagine a situation where a desired object in a photo is unintentionally obscured by other foregrounds, preventing us from obtaining its full appearance. Amodal completion~\cite{kanizsa1979organization} is a task designed for such scenarios, with the goal of reconstructing the whole object based on its visible parts. In daily life, occlusion occurs frequently, making amodal completion highly valuable in a range of downstream applications, such as autonomous vehicles and robotics.

\begin{figure*}[t!]
  \centering
  \includegraphics[width=0.95\textwidth]{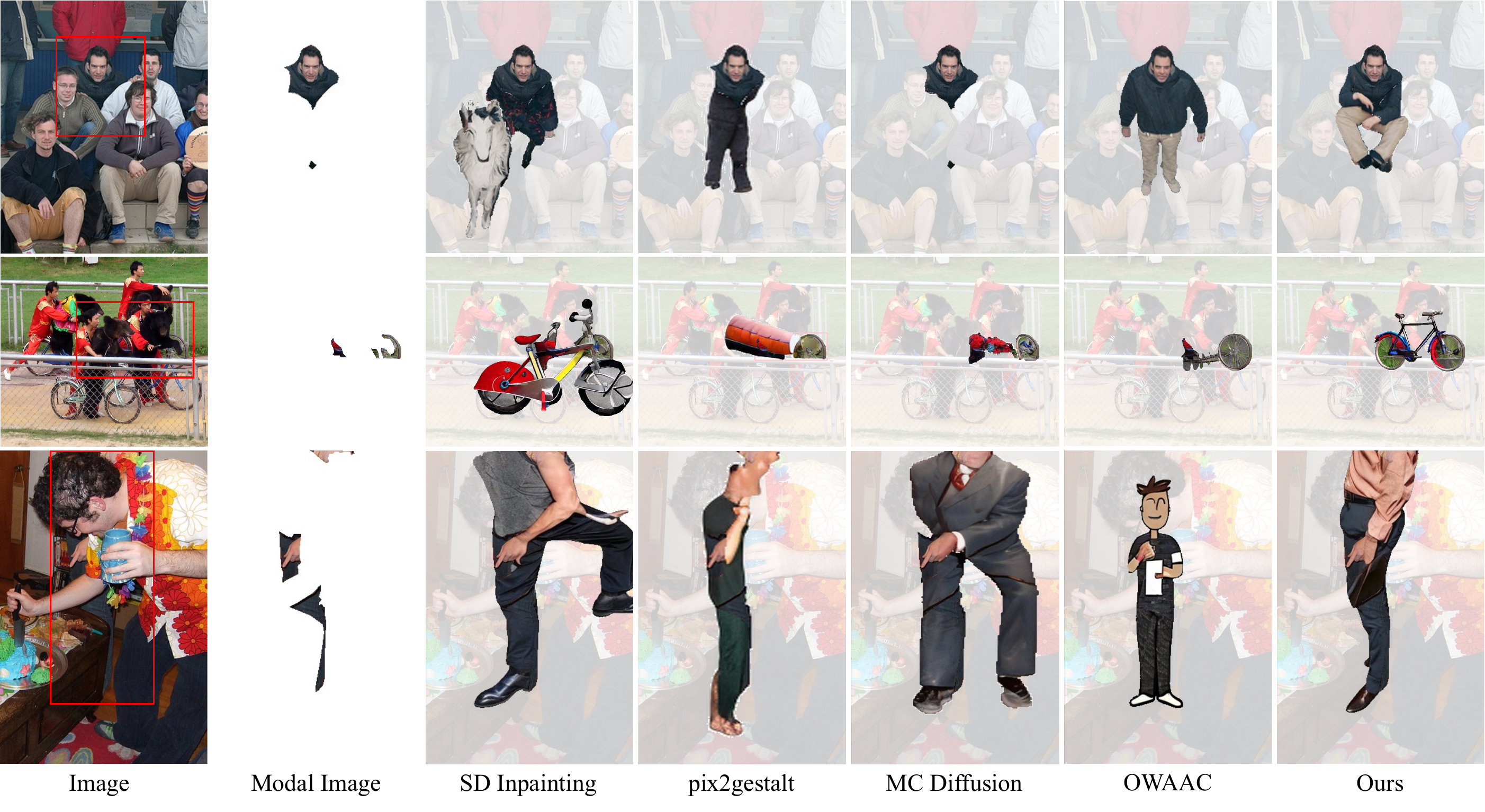}
  \vspace{-2mm}
  \caption{Completion results when meaningful parts of the target object are occluded. Stable Diffusion (SD) inpainting~\cite{rombach2022high} often generates objects other than the target object. Existing amodal completion methods~\cite{ozguroglu2024pix2gestalt,xu2024amodal,ao2025open} lack an understanding of what should be generated for the missing parts. In contrast, our method provides explicit guidance on what should be reconstructed.}
  \label{fig:baseline failure}
  \vspace{-5mm}
\end{figure*}

Similar to how humans infer the hidden parts of objects, amodal completion can greatly benefit from rich common-sense knowledge about real-world entities. However, existing methods~\cite{ozguroglu2024pix2gestalt,xu2024amodal} overlook this and heavily rely on generation capabilities of visual generative models~\cite{rombach2022high} without providing sufficient guidance about the occluded regions. MC Diffusion~\cite{xu2024amodal} proposes a training-free approach that adapts the denoising process of Stable Diffusion (SD) for amodal completion, using only category-level text prompts as guidance. pix2gestalt~\cite{ozguroglu2024pix2gestalt} fine-tunes SD on synthetic amodal completion datasets, where SD is conditioned on the input image and a modal mask. Although OWAAC~\cite{ao2025open} recently proposes to employ MLLM~\cite{lai2024lisa} in its framework, its role is restricted to segmenting the target object from abstract user queries, while SD remains conditioned solely on category-level text prompts. Consequently, these methods often produce unnatural completions in challenging occlusion scenarios and require multiple sampling attempts with different seeds to obtain satisfactory results. As illustrated in Figure \ref{fig:baseline failure}, existing methods generate an object entirely unrelated to the bicycle or fail to determine whether a person is sitting or standing based on the image context.

To overcome this limitation, we propose \textit{AmodalCG} (\textbf{Amodal C}ompletion via MLLM \textbf{G}uidance), a novel framework that effectively leverages the rich physical-world knowledge embedded in Multimodal Large Language Models (MLLMs)~\cite{llava,gpt4o,openai2023gptv,team2023gemini} for amodal completion. Specifically, AmodalCG identifies and integrates two key types of MLLM-derived guidance about the occluded content. First, the MLLM generates geometric guidance, estimating the true extent of the occluded regions. This guidance is necessary as SD becomes prone to erroneous completions when the inpainting mask is excessively larger than the actual object. To address this, AmodalCG employs MLLM to estimate the full extent of the target object and uses this prediction to resize the inpainting mask accordingly. This offers explicit cues on how much of the object should be reconstructed, preventing over-extended completion and unintended content generation. The second is semantic guidance, which provides a detailed textual description of what should be generated in the occluded area. Once the inpainting mask is resized to fit the full object, the MLLM infers the appropriate content for the occluded region. This description is then used as a text prompt for SD, giving it explicit guidance on what needs to be filled in.

However, incorporating MLLM guidance into amodal completion presents two key challenges.
First, generating MLLM guidance for every sample can be inefficient, as some cases—such as those with minimal occlusion—can already be effectively completed without detailed guidance. Second, the inherent ambiguity of amodal completion makes it difficult for MLLMs to produce accurate predictions about the hidden regions, particularly when estimating the size of the full target object.

To address these challenges, we introduce the following two strategies. First, before invoking a large-scale MLLM to generate detailed guidance, a lightweight model is used to assess the degree of occlusion and selectively triggers the large model. For samples with minimal occlusion, the framework proceeds without calling the large model, thereby reducing computational cost. Second, to alleviate the difficulty of estimating the extent of occluded regions, we adopt a multi-scale expansion strategy. Instead of producing a single estimate, the MLLM predicts multiple candidate scales for the full target object. Then, starting from the tightest prediction, our framework progressively verifies whether the target object can be fully reconstructed within each predicted region, selecting the most suitable prediction for completion. This multi-scale strategy increases the chances of the MLLM predicting an accurate object size, thereby facilitating reliable completion.

Our method enables amodal completion for open-world objects without additional training. It is simple yet highly effective, improving both amodal segmentation and occluded object recognition. Our contributions are summarized as follows:
\begin{itemize} 
\item We propose AmodalCG, a framework that selectively integrates the rich common-sense knowledge of MLLMs into challenging amodal completion.
\item We identify two key types of MLLM-derived guidance and alleviates the computational and uncertainty challenges when incorporating these guidance signals through selective large-model invocation and a multi-scale expansion strategy.
\item Our method improves amodal segmentation by 5.49\% and occluded object recognition by 2.92\% compared to the baselines, highlighting MLLMs as a promising solution for challenging amodal completion.
\end{itemize}

\begin{figure*}
  \centering
    \includegraphics[width=0.9\textwidth]{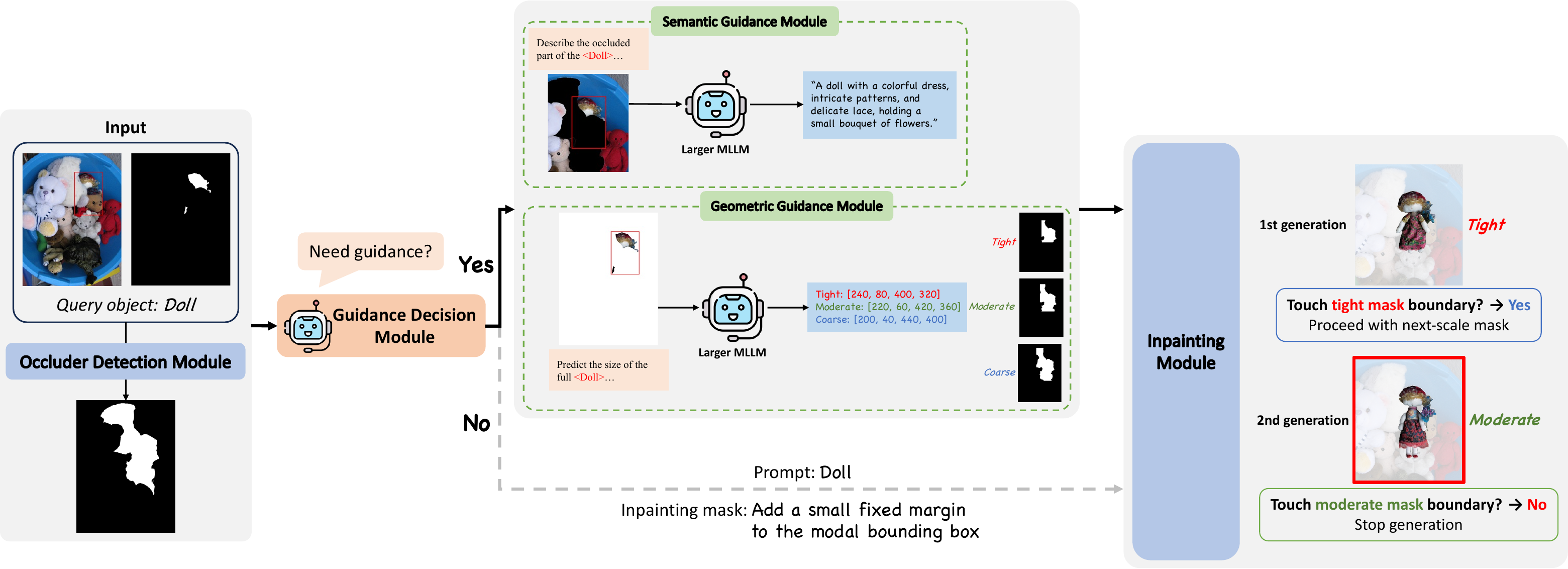}
  \caption{Overview of AmodalCG. Our framework first determines which samples would benefit from MLLM guidance (\textbf{Guidance Decision Module}). For those requiring guidance, the MLLM generates two key types of information about the occluded part of the target object: (1) the bounding box size of the full target object (\textbf{Geometric Guidance Module}) and (2) textual descriptions of the occluded region (\textbf{Semantic Guidance Module}). These are then incorporated into the completion process through a multi-scale expansion strategy, which selects the appropriate bounding box scale among the MLLM’s predictions (\textbf{Inpainting Module}).}
  \label{fig:main}
  \vspace{-3mm}
\end{figure*}

\section{Related work}
\label{sec:related work}

\noindent\textbf{Amodal Completion.}  
Early studies on amodal completion primarily focused on training task-specific models. However, this training inherently requires ground-truth appearance for the invisible regions, which are costly to obtain. Consequently, these approaches are typically trained on narrow domains such as vehicles~\cite{NEURIPS2020_bacadc62,yan2019visualizing}, humans~\cite{zhou2021human,10.1109/TMM.2022.3157036}, or indoor scenes~\cite{dhamo2019objectdrivenmultilayerscenedecomposition,ehsani2018segansegmentinggeneratinginvisible}, resulting in limited generalization to categories outside the training data.

To overcome the limitations of these datasets, recent studies~\cite{ozguroglu2024pix2gestalt,xu2024amodal,ao2025open} have leveraged the power of large-scale diffusion models, such as Stable Diffusion (SD)~\cite{rombach2022high}, which are trained on massive datasets~\cite{schuhmann2022laion}. These approaches utilize SD to directly complete the appearance of occluded objects. pix2gestalt~\cite{ozguroglu2024pix2gestalt} fine-tunes SD on synthetic datasets curated for amodal completion. MC Diffusion~\cite{xu2024amodal} bypasses the expensive fine-tuning stage by proposing a training-free pipeline that first identifies occluders from the segmentation masks~\cite{sam, liu2023grounding} and then inpaints the occluders' regions using SD. To prevent unintended generations, it clusters intermediate features of SD to retain only those features similar to the target object. Under the similar framework, OWAAC~\cite{ao2025open} enhances the completion performance by employing more accurate category-level text prompts and refining inpainting masks through morphological operations. OWAAC also employs MLLM~\cite{lai2024lisa} into its framework, whose role is restricted to segmenting the target object and does not influence the completion process. In contrast, our method incorporates the rich real-world knowledge of MLLMs directly into the generation process, providing detailed and explicit guidance to SD for completing occluded regions. 
\section{AmodalCG: Amodal Completion via MLLM Guidance}
\label{sec:method}
Given an input image $I$ along with a modal mask for the target object $M_{\mathrm{modal}}$ and, in some cases, its semantic category $P_{\mathrm{cat}}$, recent amodal completion approaches utilize visual generative models, such as Stable Diffusion (SD)~\cite{rombach2022high}, to reconstruct the occluded regions. Under the same setting, we propose AmodalCG, a framework that leverages the rich real-world knowledge of MLLMs to guide the completion process. Our method consists of five main components. First, \cref{subsec:met_occ} introduces the \textit{Occluder Detection Module}, which identifies occluders to form the inpainting mask $M_{\mathrm{inpaint}}$. Next, \cref{subsec:met_selective} describes the \textit{Guidance Decision Module}, which determines whether to invoke MLLM guidance for reconstructing the target object. We then present two modules that generate MLLM-derived information about the occluded regions: the \textit{Geometric Guidance Module} (\cref{subsec:met_mg}) and the \textit{Semantic Guidance Module} (\cref{subsec:met_pg}). Finally, \cref{subsec:met_completion} describes the \textit{Inpainting Module}, which integrates both types of MLLM guidance into the completion process through multi-scale expansion. All prompts used in our framework are provided in Appendix \cref{sec:suppple_exper}. \cref{fig:main} illustrates an overview of our pipeline.

\subsection{Occluder Detection}
\label{subsec:met_occ}
Given an input image $I$ and a modal mask $M_{\mathrm{modal}}$, the \textbf{Occluder Detection Module} outputs the inpainting mask $M_{\mathrm{inpaint}}$, defined as the union of the occluder masks~\cite{xu2024amodal,ao2025open}. To identify occluders, we first perform semantic segmentation~\cite{sam} on $I$ and then use a geometric order prediction network~\cite{lee2022instance} to determine the occlusion order of each segment.

\subsection{Selective Invocation of MLLM Guidance}
\label{subsec:met_selective}
The \textbf{Guidance Decision Module} selectively invokes MLLM guidance based on the level of occlusion. Since samples with minimal occlusion can be adequately reconstructed without additional guidance, our framework omits generating detailed MLLM guidance when the target object is regarded as minimally occluded. Unlike reasoning about occluded regions, deciding whether an object is nearly complete is a simpler task. Therefore, we employ a smaller-scale MLLM for the Guidance Decision Module to assess the necessity of MLLM guidance. Given an image of the isolated target object on a white background, the module outputs two pieces of information in JSON format: (1) a binary indicator specifying whether MLLM guidance is required, and (2) the category of the target object $P_{\mathrm{cat}}$. In \cref{subsec:met_mg,subsec:met_pg}, we describe how this information is subsequently used to generate geometric and semantic guidance.

\begin{figure}[t]
  \centering
    \includegraphics[width=\linewidth]{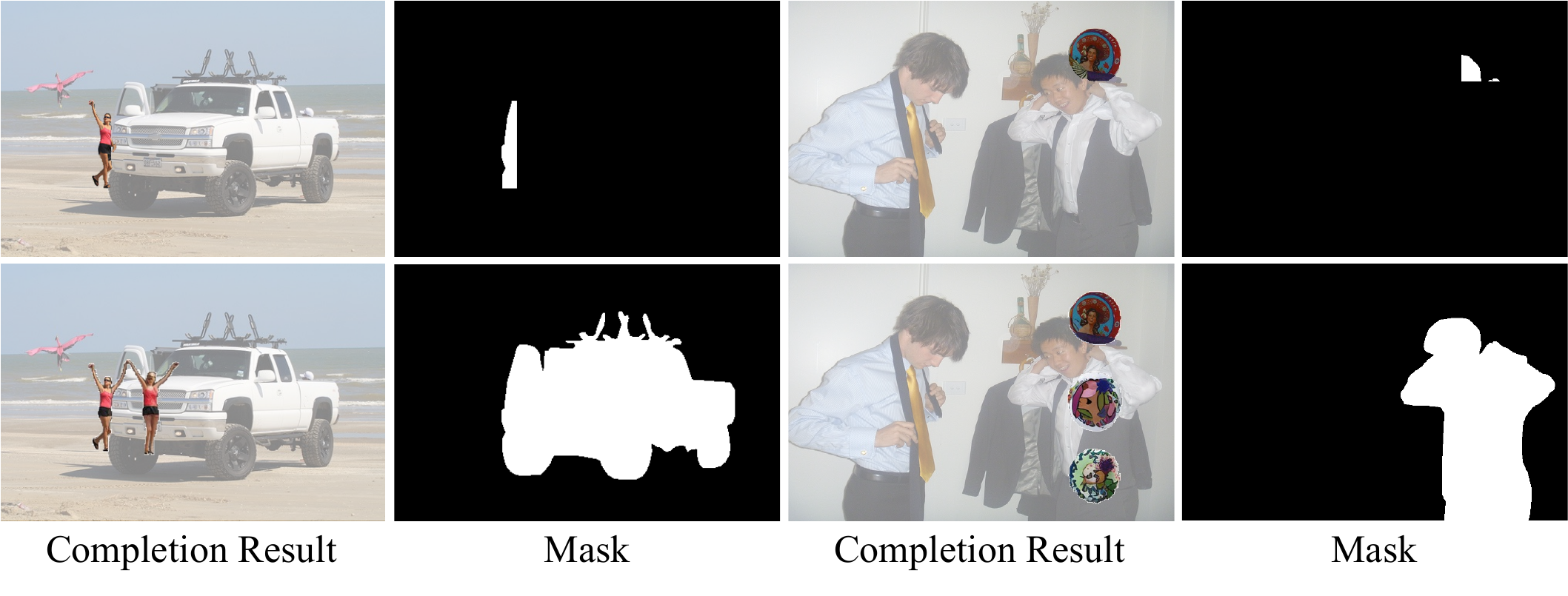}
  \caption{Amodal completion results based on inpainting mask size. Unwanted objects are generated when the inpainting mask is substantially larger than the actual occluded region.}
  \label{fig:mask_failure}
  \vspace{-6mm}
\end{figure}

\subsection{Estimating the Extent of Occluded Regions}
\label{subsec:met_mg}
One type of guidance used in our framework is geometric guidance, which represents the estimated size of the full target object, including its occluded regions. We first describe why this guidance is important for amodal completion and then details how the \textbf{Geometric Guidance Module} predicts the extent of the full target object.

\noindent\textbf{Geometric guidance is crucial to prevent unintended generation.} A major reason existing amodal completion methods often generate undesired objects beyond the target object is the use of an excessively large inpainting mask $M_{\mathrm{inpaint}}$ compared to the actual occluded area. As illustrated in \cref{fig:mask_failure}, an inpainting mask adjusted to fit the target object produces precise completions, whereas an unadjusted, overly large mask tends to generate unintended objects outside the target region. Based on this observation, the Geometric Guidance Module predicts the bounding box of the full target object and redefines the inpainting mask $M_{\mathrm{inpaint}}^*$ as the intersection between the predicted bounding box $\hat{M}_{\mathrm{bbox}}$ and the original inpainting mask $M_{\mathrm{inpaint}}$:
\begin{align}
    M_{\mathrm{inpaint}}^* = \hat{M}_{\mathrm{bbox}}\cap M_{\mathrm{inpaint}}.
\end{align}
By providing the Inpainting Module with explicit guidance on how much of the object should be generated, our method effectively suppresses unnecessary object creation and prevents over-extension of the target object.

\noindent\textbf{Estimating the extent of occluded regions.} 
This geometric guidance is crucial for all samples, regardless of occlusion level, since even minimally occluded objects may have overly large inpainting masks (see \cref{fig:mask_failure} for an example). Accordingly, we adopt two different strategies based on the output of the Guidance Decision Module. If the module determines that MLLM guidance is unnecessary (\eg, samples with minimal occlusion), the framework assumes that only minor completion is required. In this case, $\hat{M}_{\mathrm{bbox}}$ is obtained by slightly enlarging the modal bounding box with a fixed margin. Conversely, if MLLM guidance is deemed necessary, the framework assumes that extensive completion is required, and thus $\hat{M}_{\mathrm{bbox}}$ is predicted by the MLLM. However, directly predicting an accurate $\hat{M}_{\mathrm{bbox}}$ is challenging for the MLLM due to the high uncertainty inherent in occluded regions. Therefore, we jointly exploit the image generation capability of the Inpainting Module to mitigate this uncertainty. Specifically, the MLLM is instructed to predict three bounding boxes at different scales: \textit{tight}, \textit{moderate}, and \textit{coarse}. Then, the Inpainting Module progressively evaluates each bounding box, starting from the tightest prediction, until the object can be fully reconstructed within the prediction, as further described in \cref{subsec:met_completion}.

\begin{figure}[t]
  \centering
    \includegraphics[width=\linewidth]{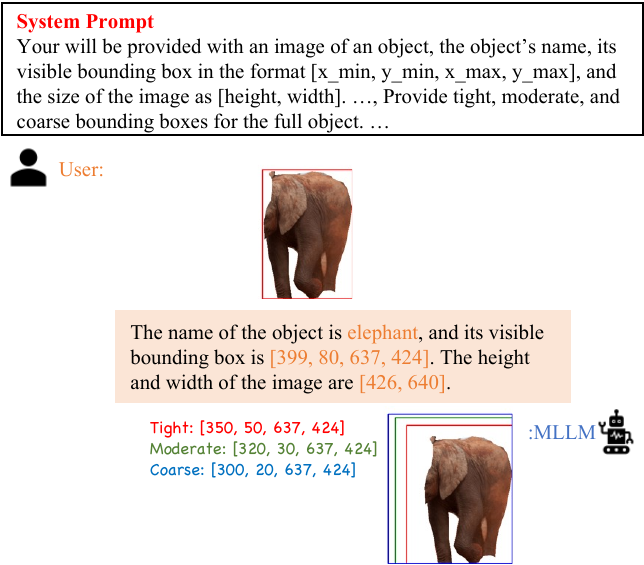}
\vspace{-2mm}
  \caption{Example of the Geometric Guidance Module predicting multi-scale bounding boxes for the full target object.}
  \label{fig:our_mg}
\end{figure}

\begin{figure}
  \centering
    \includegraphics[width=\linewidth]{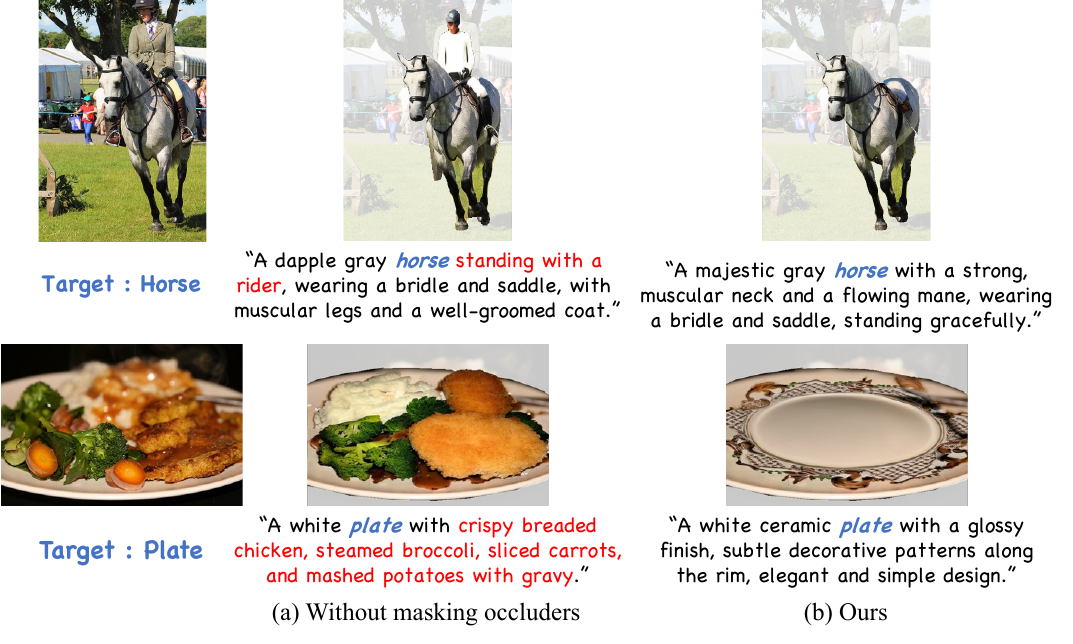}
\vspace{-5mm}
  \caption{Without masking occluders, MLLMs tend to describe occluders as well, resulting in extraneous generation in the final output. We highlight the semantic category of the target object in \textit{\textcolor{blue}{blue}} and the descriptions of the occluders in \textcolor{red}{red}.}
  \label{fig:prompt_failure}
  \vspace{-5mm}
\end{figure}

\noindent\textbf{Input Prompt for the MLLM.} We provide three types of information as a text prompt to the MLLM: (1) the coordinates of the modal bounding box, (2) the image size, and (3) the semantic category. For the visual prompt, we isolate the target object on a white background and highlight its modal bounding box in red, allowing the MLLM to clearly associate the coordinates provided in the textual prompt with the corresponding region in the image. Based on this input, the MLLM predicts \textit{tight}, \textit{moderate}, and \textit{coarse} bounding boxes of the full target object. \cref{fig:our_mg} illustrates our method, and the exact prompts used in the experiments are provided in Appendix \cref{sec:suppple_exper}.

\subsection{Generating Detailed Textual Descriptions}
\label{subsec:met_pg}
The \textbf{Semantic Guidance Module} extends beyond a category level prompt $P_{\mathrm{cat}}$ by generating $P_{\mathrm{long}}$, a detailed textual description specifying what should be generated in the occluded regions. This module is invoked only when the Guidance Decision Module determines that MLLM guidance is necessary. Below, we describe how the MLLM generates descriptions of occluded regions.
\[
P =
\begin{cases}
P_{\mathrm{long}}, & \text{if Semantic Guidance Module is invoked}, \\
P_{\mathrm{cat}}, & \text{otherwise.}
\end{cases}
\]

Unlike existing description generation methods that focus on visible parts of scenes, our focus is on generating descriptions for occluded regions of an object. However, describing the occluded parts, rather than the visible ones, presents a unique challenge. As shown in Figure \ref{fig:prompt_failure}, existing methods~\cite{chen2023shikra,jiang2024joint,yang2024fine}, which typically use visual marks to indicate the target object, often include descriptions of the occluders when they significantly overlap with the target object, leading to the generation of unintended objects in the final outputs. In the figure, although the rider and broccoli are occluders in each image, the MLLM provides descriptions about them, causing their inclusion in the final outputs. Thus, in amodal completion, it is crucial to prevent occluders from influencing the MLLM's responses.

\begin{figure}[t]
  \centering
    \includegraphics[width=\linewidth]{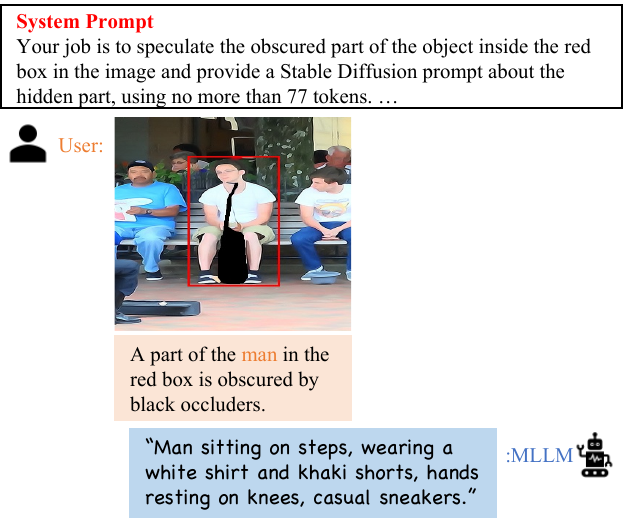}
  \caption{Example of the Semantic Guidance Module generating detailed descriptions of occluded regions.}
  \label{fig:our_pg}
  \vspace{-6mm}
\end{figure}

\begin{figure*}[t]
  \centering
    \includegraphics[width=\textwidth]{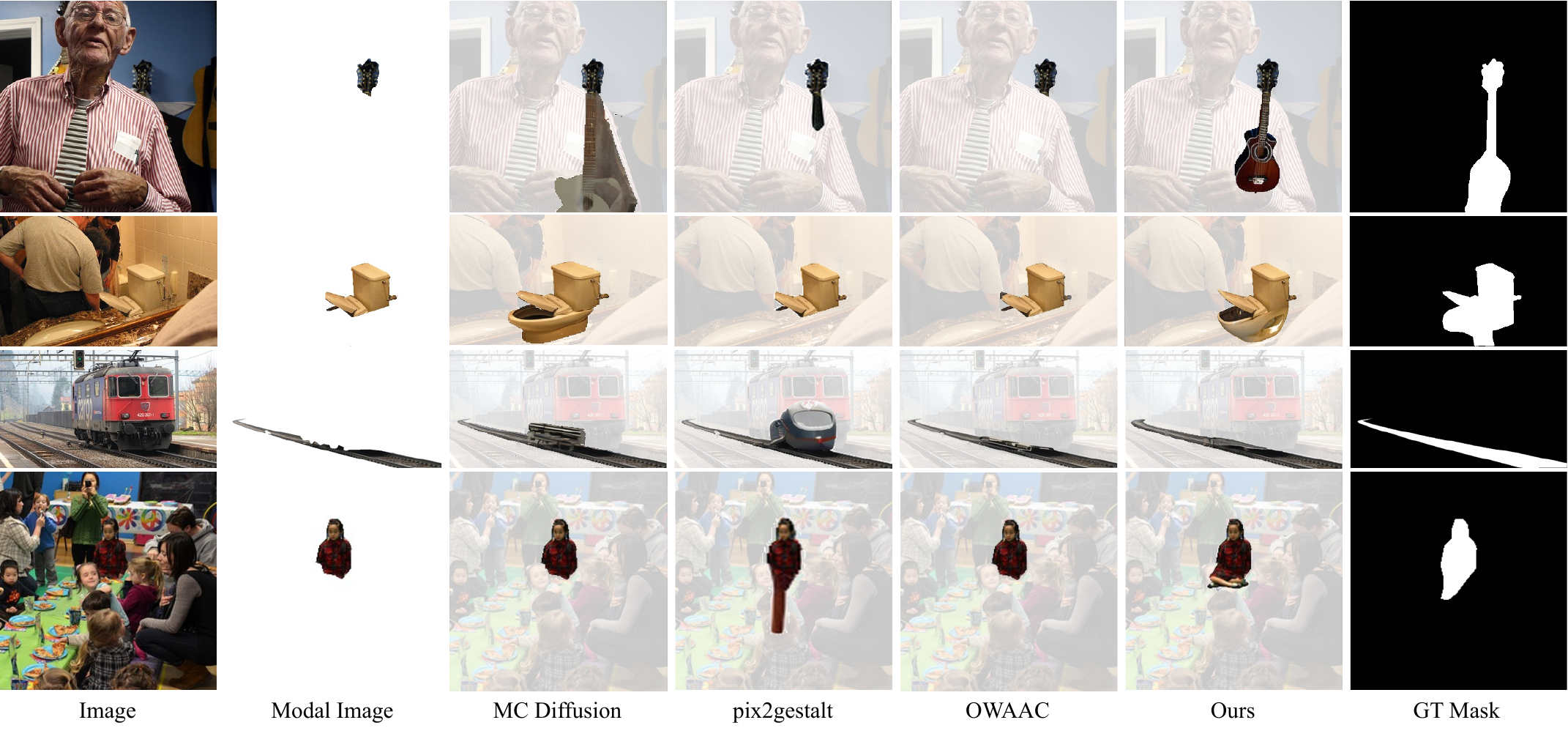}
    \vspace{-4mm}
  \caption{Qualitative evaluation of our method.}
  \label{fig:exper_seg}
\end{figure*}

Interestingly, we observe that using a visual prompt that masks the occluders effectively mitigates this issue. By removing the occluders’ appearance, this approach allows the MLLM to easily distinguish the occluders from the target object, thereby minimizing their influence on the MLLM’s response. It also allows the MLLM to understand the overall context of the image by preserving the appearance of the parts outside the occluders. Figure \ref{fig:our_pg} illustrates our method for generating descriptions of occluded regions.

\subsection{Completion with MLLM Guidance}
\label{subsec:met_completion}
Finally, the \textbf{Inpainting Module} reconstructs the appearance of the target object using the two types of guidance and outputs the completed object $\hat{I}_{\mathrm{amodal}}$ along with its segmentation mask $\hat{M}_{\mathrm{amodal}}$.
We first place the target object on a gray background $I_{\mathrm{bkgd}}$ and perform inpainting using the resized inpainting mask $M_{\mathrm{inpaint}}^{*}$ and the text prompt $P$.
After inpainting, we separate the reconstructed target object from $I_{\mathrm{bkgd}}$ by obtaining the background mask $\hat{M}_{\mathrm{bkgd}}$ using SAM~\cite{sam} and then inverting it to derive the amodal mask of the target object: $\hat{M}_{\mathrm{amodal}} = (1 -\hat{M}_{\mathrm{bkgd}})\cup M_{\mathrm{modal}}$. This approach is used because segmenting the background is generally easier, whereas the target object often contains complex internal details.

\noindent\textbf{Multi-scale expansion.} As described in \cref{subsec:met_mg}, the Geometric Guidance Module outputs three different scales of $\hat{M}_{\mathrm{bbox}}$ when MLLM guidance is invoked. In such cases, the Inpainting Module leverages its image generation capability to determine an appropriate mask scale among them. Specifically, the module starts from the tightest prediction and checks whether the target object can be fully reconstructed within the prediction. If the object is successfully reconstructed, the generated result is returned as the final output; otherwise, the module proceeds to the next larger scale and repeats the verification. This progressive process ensures that the target object is fully reconstructed within the predicted region while avoiding unintended generation. To verify whether the object has been fully reconstructed, we check whether the generated object touches the boundary of $\hat{M}_{\mathrm{bbox}}$. If the generated object reaches the boundary, we regard it as incomplete; otherwise, it is considered fully reconstructed. Although our method can continue to refine the mask by incrementally enlarging it beyond the coarsest prediction, we limit the expansion process to three scales in our experiments for efficiency.
\section{Experiments}
\label{sec:exper}
When evaluating amodal completion, two key aspects should be considered. The first is whether the target object is fully generated, and the second is whether the appearance of the object is naturally reconstructed. Following the evaluation methods of the baselines, we assess these aspects through two tasks: Amodal Segmentation and Occluded Object Recognition. We first present the performance of our method in amodal segmentation, followed by results for occluded object recognition. Finally, we validate the effectiveness of each component of our method.

\noindent\textbf{Implementation.} We use InternVL3.5-8B~\cite{wang2025internvl3} as the Guidance Decision Module and GPT-4o~\cite{gpt4o} for both the Geometric and Semantic Guidance Modules. Stable Diffusion v2 inpainting model~\cite{rombach2022high} is used for the Inpainting Module. In the Geometric Guidance Module, a 10\% margin is added to the modal bounding box when MLLM guidance is not invoked. Detailed experimental settings are provided in Appendix \cref{sec:suppple_exper}.

\begin{table*}[t]
  \centering
  \begin{tabular}{@{}lccccccccccc@{}}
    \toprule
    \multirow{2}{*}{{\textbf{Method}}} &  \multicolumn{3}{c}{COCO-A} & & \multicolumn{3}{c}{BSDS-A} & & \multicolumn{3}{c}{MP3D-A} \\
    \cline{2-4} \cline{6-8} \cline{10-12}
    & Hard & Moderate & Easy & & Hard & Moderate & Easy & & Hard & Moderate & Easy \\
    \midrule
    pix2gestalt & 64.77 & 80.32 & 85.93 & & 55.74 & 86.53 & 87.35 & & 48.81 & 70.92 & 78.74 \\
    MC Diffusion & 59.01 & 73.17 & 86.90 & & 57.37 & 64.73 & 74.59 & & 42.66 & 64.29 & 74.67 \\
    OWAAC & 51.06 & 62.41 & 78.72 & & 54.10 & 64.86 & 71.56 & & 46.76 & 66.52 & 75.75 \\
    \textbf{Ours} & \textbf{75.09} & \textbf{86.49} & \textbf{92.37} & & \textbf{67.09} & \textbf{86.60} & \textbf{90.25} & & \textbf{51.73} & \textbf{75.83} & \textbf{84.09} \\
    \bottomrule
  \end{tabular}
  \vspace{-2mm}
  \caption{Results on amodal segmentation by occlusion ratio.}
  \vspace{-5mm}
  \label{tab:seg}
\end{table*}

\subsection{Amodal Segmentation}
\label{subsec:amodal seg}
\noindent\textbf{Evaluation Details}
Amodal segmentation evaluates the similarity between the segmentation mask of the completed object and the ground truth mask of the full object using mean Intersection-over-Union (mIoU). Although amodal segmentation does not consider the appearance of the reconstructed object and multiple valid ground truth masks may exist for the occluded regions, it allows us to assess whether the target object is accurately reconstructed without incompletion or overextension by comparing the similarity of the masks. We evaluate our method on three datasets: COCO-A~\cite{cocoa}, BSDS-A~\cite{cocoa}, and MP3D-A~\cite{zhan2024amodal}. These datasets include a variety of objects commonly found in everyday life and are the most frequently used datasets for amodal segmentation. 

\noindent\textbf{Results}
Table \ref{tab:seg} shows that our method outperforms all baselines and remains the most robust under high-occlusion scenarios. Following~\cite{xu2024amodal}, we define samples with an occlusion ratio above 0.5 as hard, below 0.2 as easy, and the rest as moderate. As illustrated in Figure~\ref{fig:exper_seg}, existing methods struggle with heavily occluded objects due to their inability to incorporate detailed information about the occluded regions. In contrast, our method generates detailed guidance about occluded regions by leveraging the rich knowledge of the MLLM, thereby significantly improves amodal segmentation performance compared to the baselines. Specifically, our method can prevent the generation of extraneous objects by utilizing an inpainting mask that fits the target object's size, and can also generate more natural reconstructions through progressive mask expansion guided by detailed prompts. The effectiveness of each component of our method is further explored in \cref{subsec:ablation}.

\subsection{Occluded Object Recognition}
\label{subsec:OOR}
\noindent\textbf{Evaluation Details}
Occluded object recognition is a classification task that classifies the class of an occluded object. This task allows us to evaluate how well the appearance of the occluded object has been restored. Following the setting used in pix2gestalt, we employ CLIP~\cite{clip} as the classification model and evaluate objects placed on a white background. For the dataset, we utilize the Occluded and Separated COCO datasets~\cite{zhan2210tri}, which contain 80 COCO semantic categories. The occluded COCO dataset consists of occluded objects represented as a single segment, while the separated COCO dataset consists of occluded objects represented as multiple segments, making it more challenging.

\begin{table}
  \centering
  \begin{tabular}{@{}lcccc@{}}
    \toprule
    \multirow{2}{*}{\textbf{Method}} & \multicolumn{2}{c}{Occluded} & \multicolumn{2}{c}{Separated} \\
    \cline{2-3} \cline{4-5} 
    & Top 1 $\uparrow$ & Top 3 $\uparrow$ & Top 1  $\uparrow$ & Top 3 $\uparrow$ \\
    \midrule
    No completion & 34.00 & 49.26 & 21.10 & 34.70 \\
    pix2gestalt & 43.39 & 58.97 & 31.15 & 45.77 \\
    MC Diffusion & 44.74 & 62.07 & 34.50 & 49.72 \\
    OWAAC & 40.50 & 55.30 & 27.83 & 40.97 \\
    \textbf{Ours} & \textbf{45.06} & \textbf{62.99} & \textbf{40.01} & \textbf{56.70}\\
    \bottomrule
  \end{tabular}
  \caption{Qualitative evaluation on occluded object recognition using two datasets. We report Top 1 and Top 3 accuracy (\%) using CLIP as a classification model.}
  \vspace{-4mm}
  \label{tab:OOR}
\end{table}

\noindent\textbf{Results}
Table \ref{tab:OOR} demonstrates that our method is highly effective in completing the appearance of occluded objects compared to the baselines. As shown in the table, our method consistently outperforms existing approaches. This is because existing methods heavily rely on the prior knowledge of Stable Diffusion to reconstruct occluded objects without providing detailed guidance on what to generate for the occluded parts. In contrast, our approach leverages detailed information about the occluded regions generated by the MLLM to help the reconstruction of the occluded object, and our experimental results indicate that this approach is highly effective for amodal completion. 
Furthermore, our method performs well on the more challenging separated COCO dataset, which demonstrates that our method is also effective in difficult samples.

\begin{table}[h]
  \centering
  \begin{tabular}{@{}cc|cc|cc@{}}
    \toprule
    \multicolumn{2}{c}{COCO-A} & \multicolumn{2}{c}{BSDS-A} & \multicolumn{2}{c}{MP3D-A} \\
    \cmidrule{1-6}
    GCR $\uparrow$ & GSR $\uparrow$ & GCR $\uparrow$ & GSR $\uparrow$ & GCR $\uparrow$ & GSR $\uparrow$  \\
    \midrule
    94.18 & 48.73 & 98.17 & 35.02 & 99.62 & - \\
    \bottomrule
  \end{tabular}
  \caption{Guidance Call/Skip Rate (GCR/GSR) (\%) of the Guidance Decision Module. GSR is not reported for MP3D-A, as it does not contain samples with low occlusion (\ie, $<$10\%).}
  \label{tab:ab_gdm}
\end{table}

\subsection{Ablation Study}
\label{subsec:ablation}
\begin{figure}
    \centering
    \includegraphics[width=\linewidth]{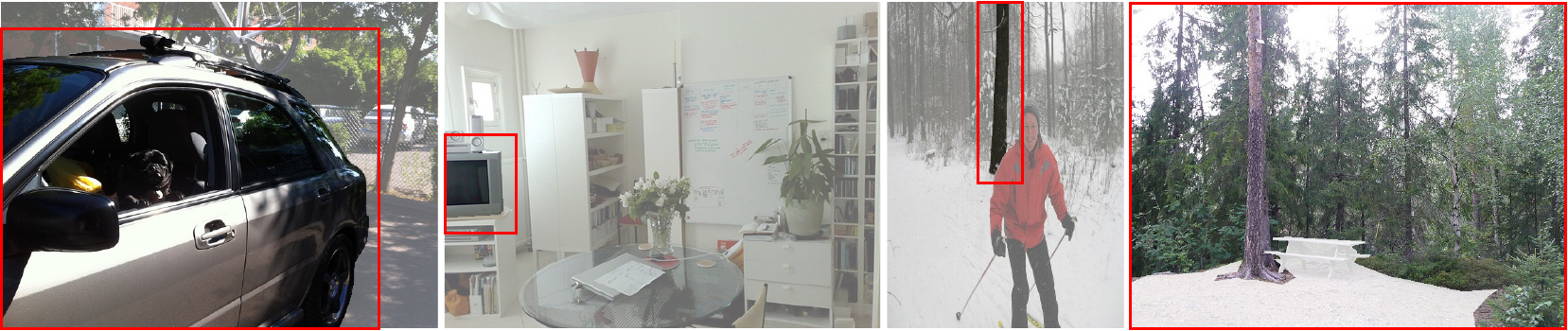}
    \caption{Failure cases where the Guidance Decision Module failed to skip guidance.}
    \label{fig:gdm_failure}
\end{figure}

\noindent\textbf{Analysis of Guidance Decision Module} We first evaluate whether the Guidance Decision Module appropriately invokes MLLM guidance when needed. The module is assessed in two aspects: (1) whether it correctly invokes guidance — \textbf{Guidance Call Rate (GCR)}, and
(2) whether it appropriately avoids unnecessary guidance — \textbf{Guidance Skip Rate (GSR)}. Since ground-truth labels indicating whether guidance is required are unavailable, we approximate them by defining samples with an occlusion ratio greater than 50\% as those that truly require guidance, and samples with less than 10\% occlusion as those that likely do not.
As shown in \cref{tab:ab_gdm}, our module accurately invokes guidance for samples that truly require it, showing high GCR. In contrast, GSR is relatively lower. We found that this is mainly due to cases, such as those shown in \cref{fig:gdm_failure}, where the object is truncated by the image boundary or belongs to ambiguous background regions (\ie, trees). Since the occlusion ratio is computed within the image boundary, these objects exhibit low occlusion ratios despite being incomplete, which causes the module to invoke guidance even when the ratio is low.

\begin{figure}[b]
  \centering
    \includegraphics[width=\linewidth]{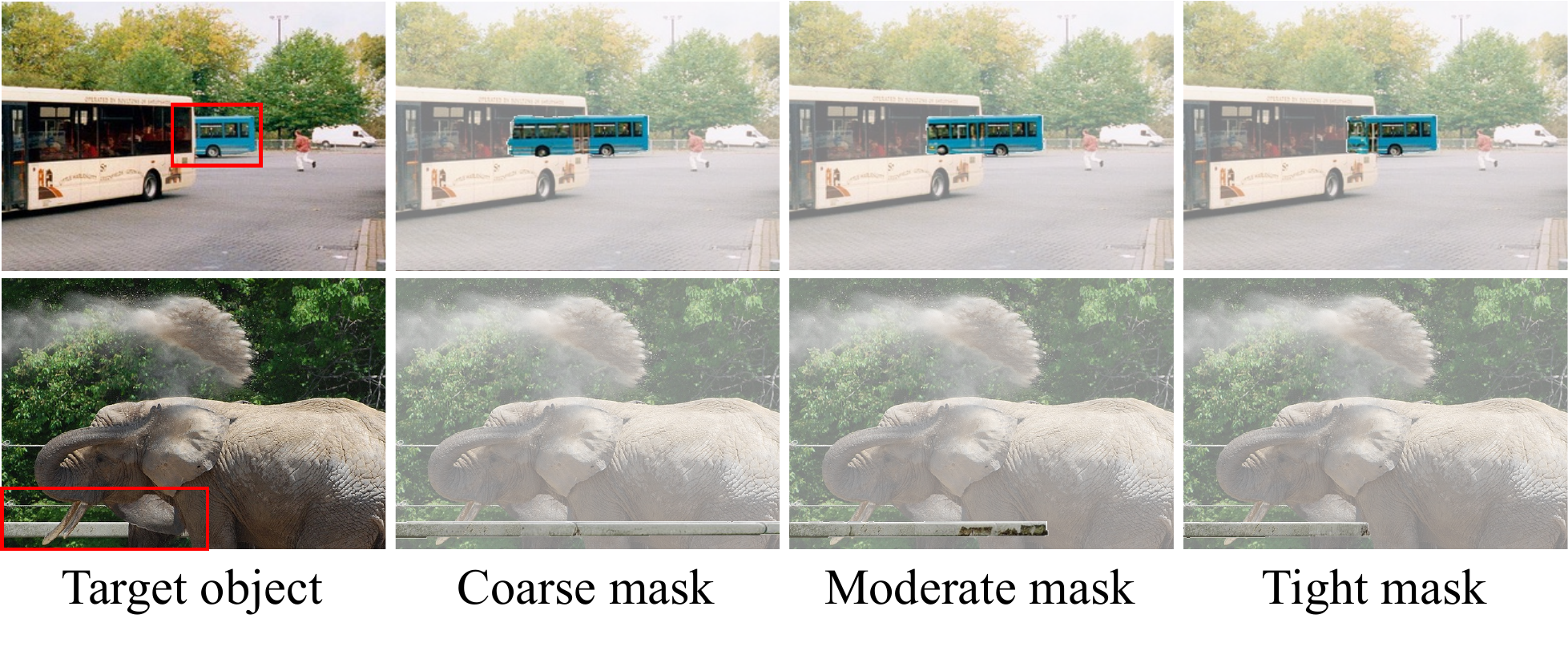}
  \caption{Completion results across different mask scales.}
  \label{fig:mask_diversity}
  \vspace{-3mm}
\end{figure}

\noindent\textbf{Analysis of Geometric Guidance Module}
Our geometric guidance, which adjusts the inpainting mask to align with the actual size of the target object, plays a key role in performance improvement, as applying it alone yields strong results, as illustrated in Table~\ref{tab:ab_ours}. We attribute this to the frequent occurrence of substantial occluders in natural scenes, which causes the inpainting mask to become overly large and leads to extraneous generation. Additionally, we examine how completion results can be refined by our multi-scale expansion strategy in \cref{fig:mask_diversity}. As illustrated, our method enables the controllable generation of objects by utilizing masks of different scales.

\begin{table}
  \centering
  \begin{tabular}{@{}lc@{}}
    \toprule
    \multirow{2}{*}{{\textbf{Method}}} & COCO-A \\
    \cline{2-2}
    & mIoU $\uparrow$ (\%) \\
    \midrule
    SD Inpainting (w/o guidance) & 74.77 \\
    + geometric guidance (single scale) & 85.64 \\
    + semantic guidance & 85.60 \\
    + multi-scale expansion \textbf{(Ours)} & \textbf{86.31} \\
    \bottomrule
  \end{tabular}
  \caption{Ablation study of our framework. Each component is sequentially added.}
  \label{tab:ab_ours}
  \vspace{-2mm}
\end{table}

\begin{figure}[tbp]
    \centering
    \includegraphics[width=\linewidth]{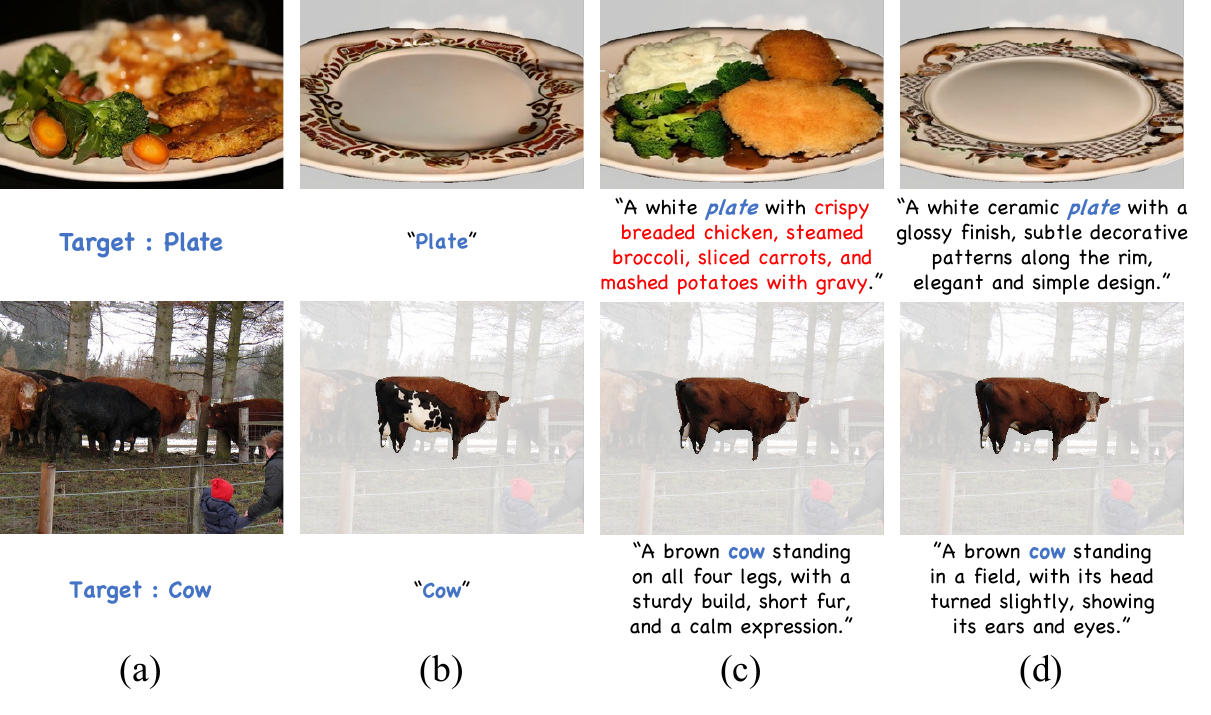}
    \vspace{-20pt}
    \caption{(a) Image of the target object. (b) Completion results with a semantic category as a text prompt. (c) Completion results with a description generated without masking occluders. (d) Ours.}
    \label{fig:ab_prompt}
    \vspace{-15pt}
\end{figure}

\noindent\textbf{Analysis of Semantic Guidance Module} 
After resizing the inpainting mask to match the target object’s size, our detailed descriptions from the Semantic Guidance Module helps restore the object with a plausible pose and appearance. As shown in \cref{fig:ab_prompt}, these descriptions effectively capture the pose and visual characteristics of the hidden regions, enabling the generation of realistic appearances such as the cow’s patterns or a clean plate.  However, the effect of semantic guidance on COCO-A is minimal, as illustrated in \cref{tab:ab_ours}. We attribute this to the fact that COCO-A primarily consists of objects with limited pose variation, such as static items or natural backgrounds, and that the segmentation task itself does not account for appearance. In fact, our semantic guidance proves beneficial for occluded object recognition, as shown in \cref{tab:OOR_ab_prompt}. We present further analysis in Appendix \cref{sec:supple_prompt}.

\begin{table}[h]
  \centering
  \resizebox{\linewidth}{!}{
  \begin{tabular}{@{}lcccc@{}}
    \toprule
    \multirow{2}{*}{\textbf{Method}} & \multicolumn{2}{c}{Occluded} & \multicolumn{2}{c}{Separated} \\
    \cline{2-3} \cline{4-5} 
    & Top 1 $\uparrow$ & Top 3 $\uparrow$ & Top 1  $\uparrow$ & Top 3 $\uparrow$ \\
    \midrule
    Ours w/o semantic guidance & 44.54 & 62.65 & 39.04 & 56.33 \\
    \textbf{Ours} & \textbf{45.06} & \textbf{62.99} & \textbf{40.01} & \textbf{56.70}\\
    \bottomrule
  \end{tabular}
  } 
  \caption{Effectiveness of our semantic guidance in OOR.}
  \vspace{-4mm}
  \label{tab:OOR_ab_prompt}
\end{table}

\section{Conclusion}
In this paper, we proposed AmodalCG, a framework that selectively harnesses the rich real-world knowledge of MLLMs to guide amodal completion. First, the Guidance Decision Module selectively invoked MLLM guidance by assessing the level of occlusion. For samples requiring guidance, our framework generated two key types of guidance from MLLMs to improve the completion process. First, geometric guidance provided cues on how much of the object should be reconstructed, thus preventing extraneous generation. Second, semantic guidance offered detailed instruction about what should be generated. During the completion process, we exploited the strengths of both MLLMs and visual generative models through a multi-scale expansion strategy. Experimental results demonstrated that MLLMs can effectively enhance amodal completion, offering a promising direction for integrating large multimodal reasoning into amodal completion.

{
    \small
    \bibliographystyle{ieeenat_fullname}
    \bibliography{main}
}
\clearpage
\setcounter{page}{1}
\maketitlesupplementary

\section{Implementation and Evaluation Details}
\label{sec:suppple_exper}
In this section, we provide the implementation details of our experiments.

\subsection{Guidance Decision Module}
\noindent\textbf{Visual Prompt} We segment the visible parts of the target object and place them on a white background. Then, we crop the image with a 100-pixel margin around the target object to create the visual prompt.

\noindent\textbf{Text Prompt} We present the exact prompt used in the Guidance Decision Module, which produces two outputs: (1) whether MLLM guidance is required, and (2) the semantic category of the target object.
\begin{promptblock}
You are an expert visual annotator. You are given an image where a red bounding box highlights a target subject. Follow these instructions carefully and output only a valid JSON object.
\newline\newline
Instructions:

1. Determine amodal completion requirement:

   - Output "no" only if you are confident that the target object in the red box is already complete, minimally occluded, or cannot be further extended.
   
   - Otherwise, output "yes".
   
   - If you are uncertain, output "yes".
\newline\newline
2. Identify the category name:

   - Output the category name of the target subject.
\newline\newline
3. Do not include explanations, reasoning, or any text outside of the JSON.
\newline\newline
Output format:

\{

  "requires\_extensive\_completion": "yes" | "no",
  
  "category": str
  
\}
\newline\newline
Example output:

\{

  "requires\_extensive\_completion": "yes",
  
  "category": "Bear"
  
\}
\end{promptblock}

\subsection{Geometric Guidance Module}
\noindent\textbf{Visual Prompt} We segment the visible parts of the target object and place them on a white background. We use the full-sized image as a visual prompt, allowing the MLLM to consider the original image scale when predicting the size of the full object.

\noindent\textbf{Text Prompt} The user prompt is described in \cref{fig:our_mg}. Below is the exact system prompt we used to make the MLLM predicts the size of the full target object.

\begin{promptblock}
\textbf{System Prompt}

You will be provided with an image of an object, the object's name, its visible bounding box in the format $[x_{min}, y_{min}, x_{max}, y_{max}]$, and the size of the image as $[$height, width$]$. The bounding box is marked with a red box in the image. The object in the red box is partially obscured and your task is to estimate three bounding boxes for the entire object, including both the visible and invisible parts. Provide tight, moderate, and coarse bounding boxes for the full object in the format $[x_{min}, y_{min}, x_{max}, y_{max}]$. The tight bounding box should include minimal margin around the visible parts of the object. Just provide the three bounding boxes, without explanation.
\end{promptblock}

\subsection{Semantic Guidance Module}
\noindent\textbf{Visual Prompt} We mark the visible parts of the target object with a bounding box and mask the occluders. Then, we crop the image with a 100-pixel margin around the target object to create the visual prompt. This encourages the MLLM to focus on the surrounding regions of the target object when generating descriptions.

\noindent\textbf{Text Prompt} The user prompt is described in \cref{fig:our_pg}. For the system prompt, we assign the MLLM the task of inferring the occluded parts and instruct it to describe only the target object without including any descriptions of the occluders and background. Below is the exact system prompt we used to generate descriptions of occluded regions.
\begin{promptblock}
\textbf{System Prompt}

Your job is to speculate the obscured part of the object inside the red box in the image and provide a Stable Diffusion prompt about the hidden part, using no more than 77 tokens. Do not include the names of the occluders and descriptions of the background in the prompt, focusing solely on the object. Your response should only contain the prompt. Your response should begin with a prefix that says ‘Prompt:’.
\end{promptblock}

\subsection{Evaluation Details}
Given the inherent ambiguity of amodal completion, where multiple plausible answers can exist, previous approaches~\cite{xu2024amodal,ao2025open} have primarily relied on user studies rather than quantitative metrics, or have evaluated similarity with incomplete, occluded objects~\cite{ao2025open}, which is not robust when objects are heavily occluded. Therefore, to reliably evaluate whether the occluded regions are properly reconstructed, we adopt amodal segmentation and occluded object recognition as our main quantitative evaluation metrics, following the evaluation setting of pix2gestalt~\cite{ozguroglu2024pix2gestalt}. Amodal segmentation assesses reconstruction quality by comparing the mask of the reconstructed object with human-annotated or 3D-projected ground-truth masks. Occluded object recognition provides a complementary evaluation by checking whether the reconstructed appearance of the occluded region is consistent with the correct semantic category of the object, offering a robust measure of appearance fidelity.

We clarify our evaluation protocol. Unlike other baselines that perform multiple refinement steps through iterative generation, pix2gestalt completes the task in a single forward generation process without refinement. Therefore, to account for the inherent uncertainty of amodal completion and to ensure a fair comparison in computational cost, we evaluate pix2gestalt by generating three samples for each input and reporting their average performance.

\begin{figure}[b]
    \centering
    \includegraphics[width=0.97\columnwidth]{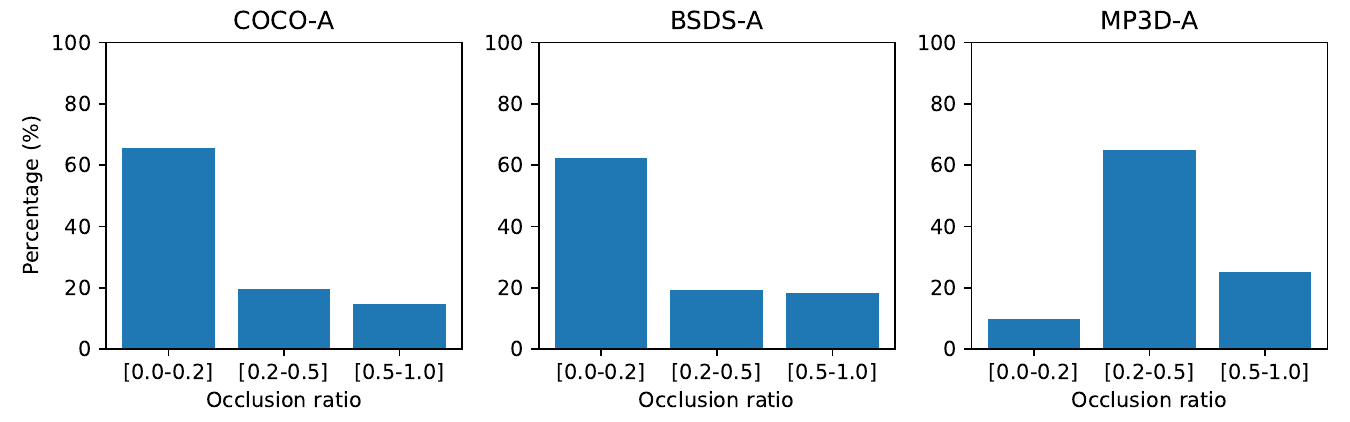}
  \caption{Occlusion percentage of the three datasets.}
  \label{fig:rebut_dataset}
\end{figure}

\begin{figure}[h]
    \centering
    \includegraphics[width=\columnwidth]{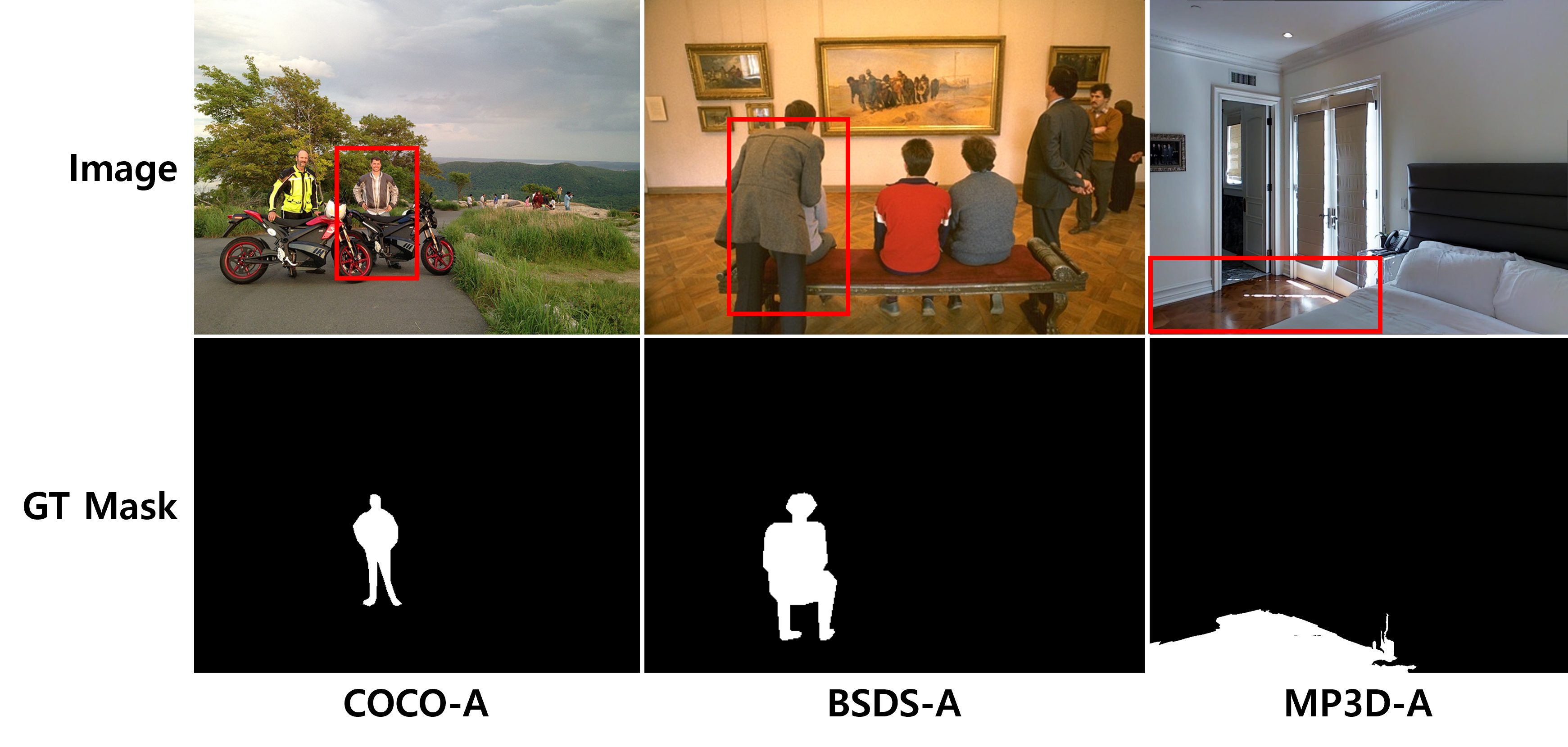}
  \caption{Examples of the three datasets.}
  \label{fig:dataset_ex}
\end{figure}

\section{Dataset Analysis}
\cref{fig:rebut_dataset} shows occlusion statistics for each dataset. All datasets consist of real images, each accompanied by a single annotated ground-truth mask. COCO-A and BSDS-A include human-annotated masks, while MP3D-A relies on 3D projection, which introduces slight noise (see \cref{fig:dataset_ex}). High-occlusion samples are less common in COCO-A and BSDS-A due to the difficulty of manual annotation. These datasets are widely used to evaluate real-world applicability, as they are the \textbf{only} real-world datasets covering numerous categories~\cite{zhan2024amodal}. 

\begin{figure*}
    \includegraphics[width=\linewidth, height=0.95\textheight]{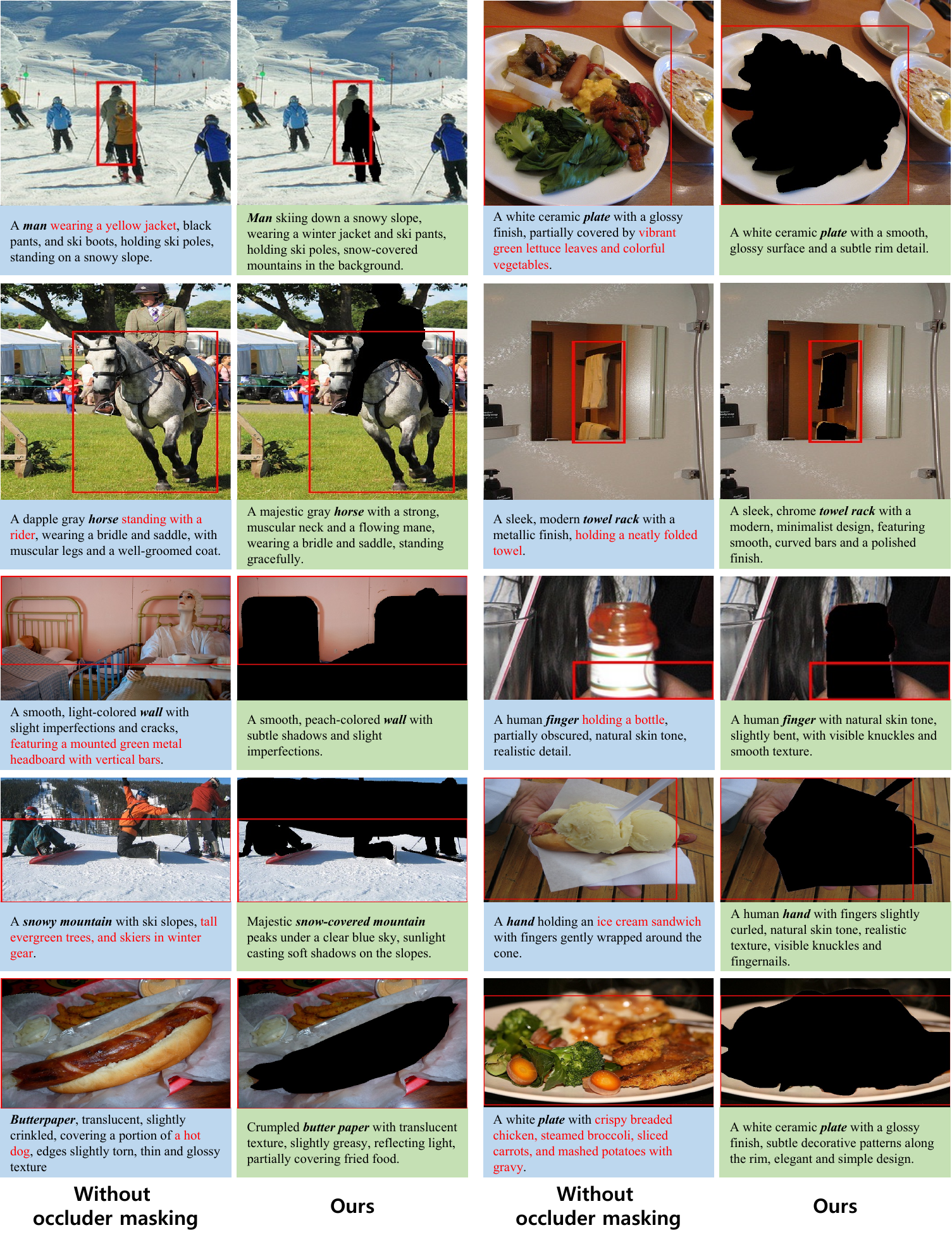}
    \caption{A comparison of the descriptions generated by the MLLM when using visual prompts with and without occluder masking. In the descriptions, the semantic category of the target object is highlighted in \textbf{\textit{bold}}, and descriptions of the occluders are highlighted in \textcolor{red}{red}.}
    \label{fig:supple_descrip_1}
\end{figure*}

\begin{figure*}[t]
    \includegraphics[width=\textwidth]{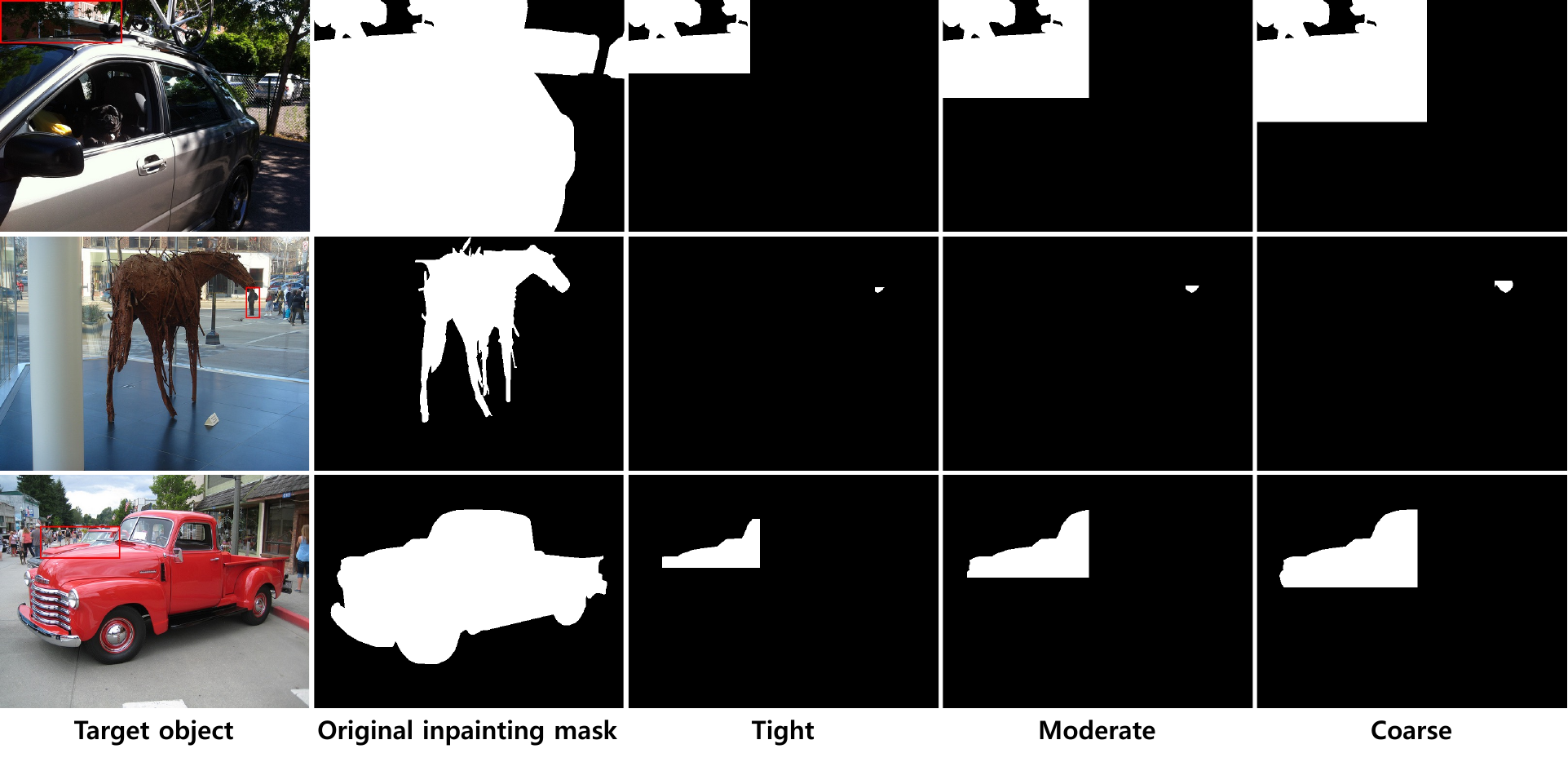}
    \vspace{-6mm}
    \caption{Examples of inpainting masks generated by our method. The target object is highlighted with a red box in each image. The target objects are a house in the first image, a man in the second, and a car in the third.}
    \label{fig:supple_mask}
\end{figure*}

\begin{figure}
    \centering
    \includegraphics[width=\linewidth,]{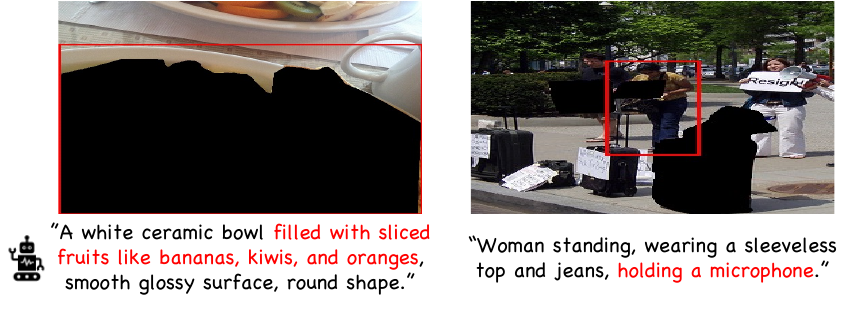}
    \vspace{-3mm}
    \caption{MLLM occasionally includes descriptions of objects that are not present in the image but are related to the image context.}
    \label{fig:limit}
\end{figure}

\section{Analysis on the Semantic Guidance Module}
\label{sec:supple_prompt}
In this section, we explore the effectiveness of the visual prompt used in the semantic guidance module and discuss why naive prompting falls short in amodal completion. Lastly, we present failure cases of the semantic guidance module.

To effectively guide amodal completion, descriptions of the occluded parts should satisfy two key conditions. First, they should exclude descriptions of the occluders, as including such descriptions may lead to the generation of unintended objects. Second, the descriptions must be consistent with the image context; otherwise, the target object may be unnaturally restored.

Masking occluders in the visual prompt enables the MLLM to generate descriptions that meet both conditions. We believe this approach helps the MLLM clearly distinguish the target object from the occluders, while the preserved shape of the occluders aids in inferring a plausible pose. To demonstrate this, we compare the MLLM's responses when using visual prompts with and without occluder masking. As shown in the \cref{fig:supple_descrip_1}, without occluder masking, the MLLM often fails to distinguish the target object from the occluders, resulting in descriptions that include details about the occluders. In contrast, our visual prompt enables the MLLM to accurately distinguish the target object, even when the target object and the occluders overlap significantly. Furthermore, the descriptions generated using our visual prompt align well with the image context.

\subsection{Analysis on failure cases} 
\label{subsec:supple_descript_fail}
We observe that our method is not entirely free from the co-occurrence bias of MLLMs. As shown in \cref{fig:limit}, while our method effectively enables the MLLM to distinguish between the target object and the occluder, it occasionally includes descriptions of co-occurring objects. For instance, in the figure, the MLLM mentions sliced fruits or a microphone, even though they are not present in the image.

\section{Analysis on the Geometric Guidance Module}
\label{sec:supple_mask}
In this section, we further analyze our approach to predicting the full target object size using the MLLM. First, we present examples of the inpainting masks generated by our method. Second, we evaluate the effectiveness of our method in predicting the true extent of the object. Finally, we show some failure cases of the geometric guidance module.

\subsection{Examples of our inpainting mask}
\cref{fig:supple_mask} shows examples of inpainting masks generated by our method. As illustrated in the figure, our approach incorporates three different scales of masks, each reflecting the characteristics of the target object. For instance, small inpainting masks are generated when only the man's head is occluded, whereas larger inpainting masks are created when substantial portions of the car or the house are occluded. This demonstrates that our method produces reasonable size estimations and effectively prevents the use of excessively large inpainting masks by adjusting the mask size to match the size of the full target object.

\begin{table}
  \centering
  \begin{tabular}{@{}lcc@{}}
    \toprule
    \multirow{2}{*}{{\textbf{Method}}} & COCO-A & BSDS-A \\
    \cline{2-3}
    & mIoU $\uparrow$ (\%) & mIoU $\uparrow$ (\%) \\
    \midrule
    Original inpainting mask & 48.00 & 52.02 \\
    \textbf{Our inpainting mask} & \textbf{78.97} & \textbf{74.60} \\
    \bottomrule
  \end{tabular}
  \caption{Accuracy of our inpainting mask.}
  \label{tab:supple_mask_acc}
  \vspace{-3mm}
\end{table}

\subsection{Effectiveness of the Geometric Guidance Module}
\label{subsec:supple_mask_acc}
To evaluate the capability of our method in estimating the target object's size, we assess the accuracy of our resized inpainting mask derived from the predicted bounding box for the full target object. We compare the similarity between the inpainting mask adjusted using the ground truth bounding box of the full target object and the mask adjusted by our method by computing mIoU. As shown in \cref{tab:supple_mask_acc}, our method leverages geometric guidance to generate inpainting masks that match the target object, effectively preventing the use of overly large masks.

\begin{figure}
    \centering
    \includegraphics[width=\linewidth]{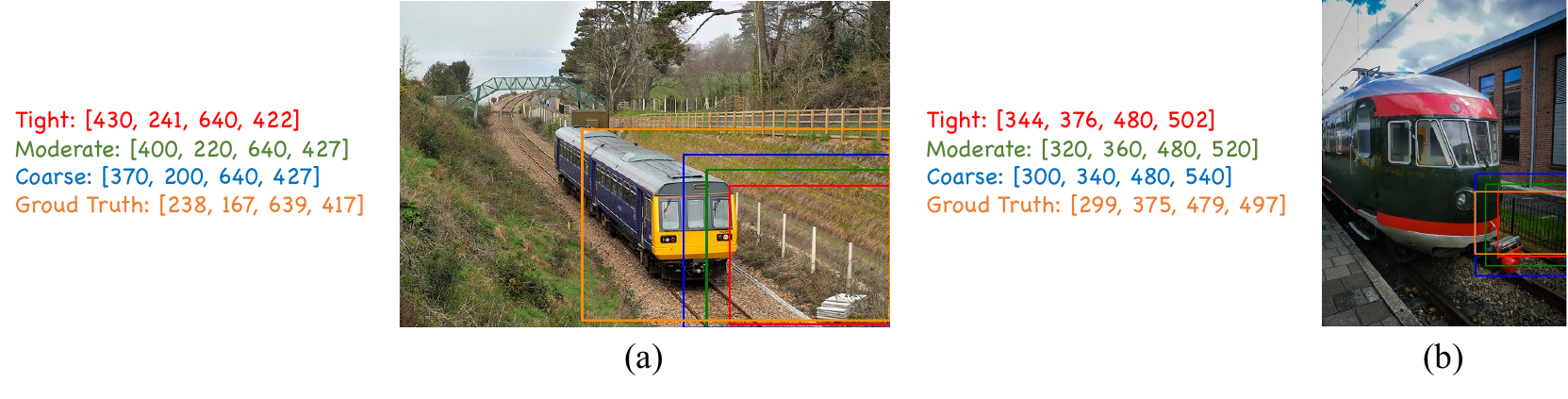}
    \caption{Examples of failure cases. Our method tends to avoid overly aggressive estimation, even in the case of \textcolor{blue}{coarse} estimation. (a) Fails to predict the size of a long fence. (b) Matches the \textcolor{orange}{ground truth size} well but still fails to predict a long fence.}
    \label{fig:supple_mask_fail}
\end{figure}

\subsection{Analysis on failure cases}
Although our method produces reasonable estimations in most cases, as previously discussed, it sometimes avoids excessively aggressive estimation, even in the case of coarse estimation. As shown in \cref{fig:supple_mask_fail}, although the fences could be long, our method predicts only the size of short fences. Nevertheless, this limitation can be alleviated by incorporating more scales in the multi-scale expansion, enabling further enlargement of the mask at the expense of efficiency.

\section{Additional Qualitative Results of Amodal Completion}
\begin{figure*}
    \includegraphics[width=\textwidth, height=0.9\textheight]{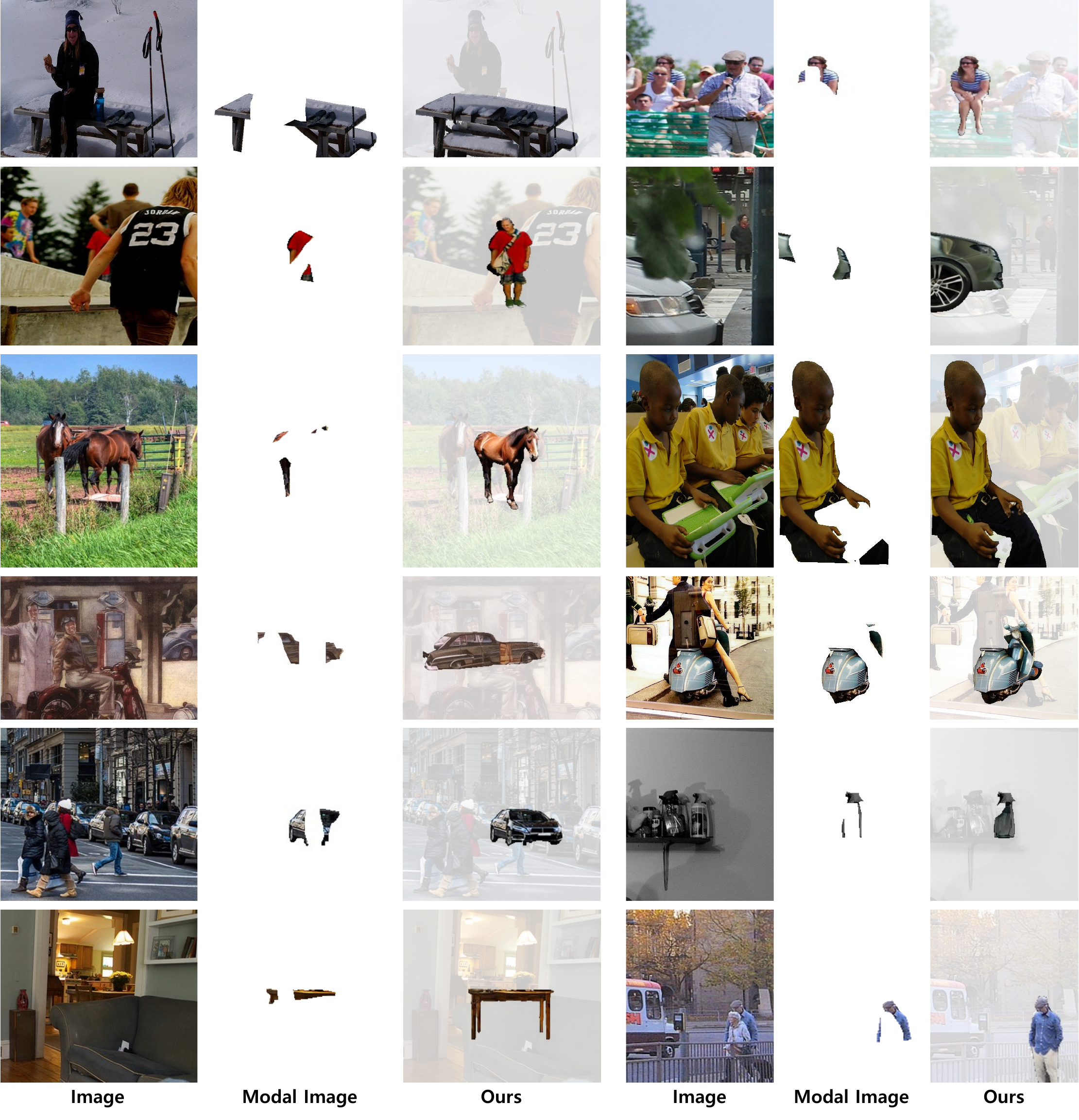}
    \caption{Additional qualitative results of our method in amodal completion.}
    \label{fig:supple_add3}
\end{figure*}

\begin{figure*}
    \includegraphics[width=\textwidth, height=0.9\textheight]{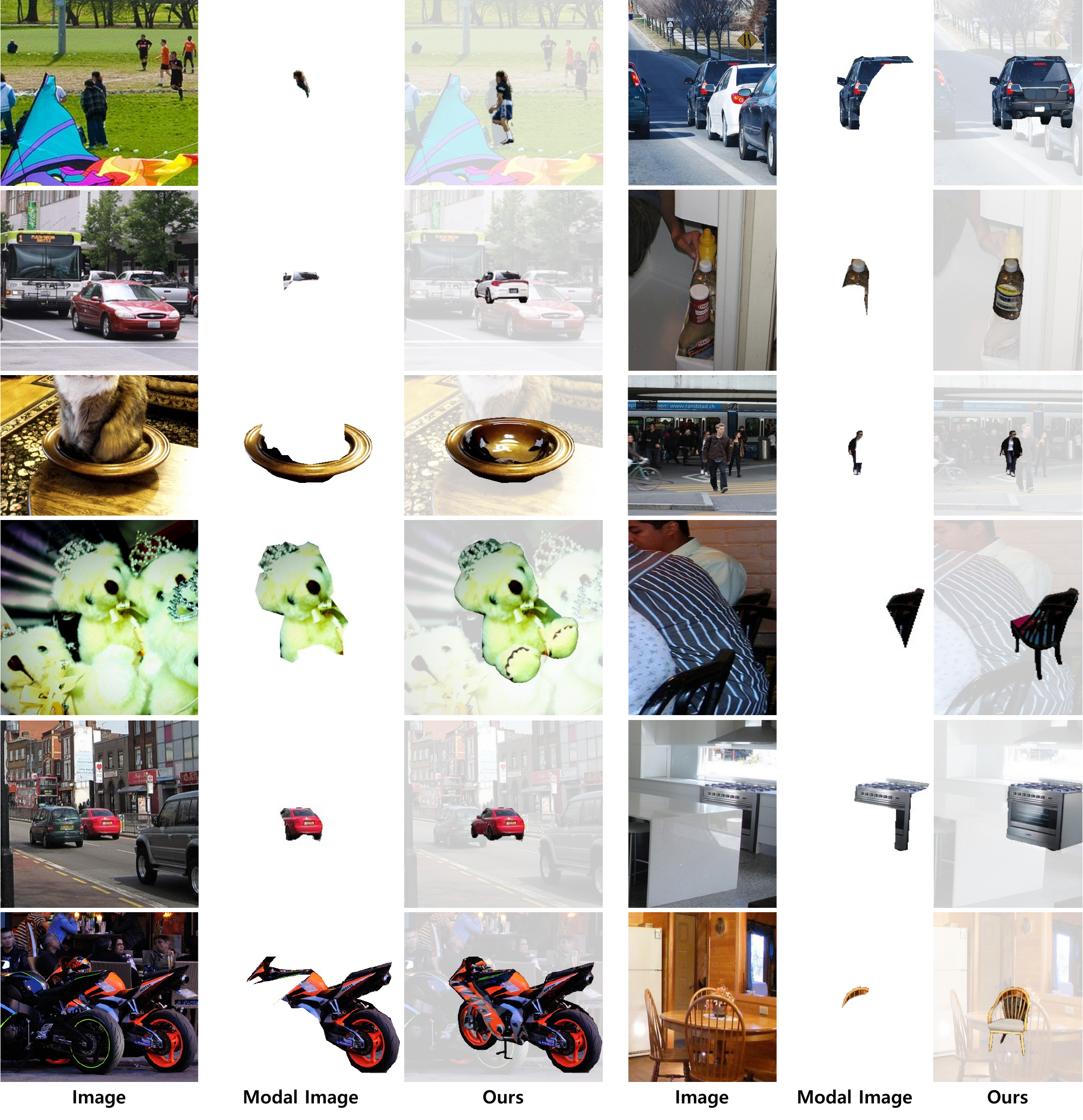}
    \caption{Additional qualitative results of our method in amodal completion.}
    \label{fig:supple_add4}
\end{figure*}

In this section, we present additional qualitative results (\cref{fig:supple_add3,fig:supple_add4}) of our method in amodal completion.

\end{document}